\documentclass[letterpaper,10pt,conference]{ieeeconf}
\usepackage{amsmath,amsfonts}
\usepackage{algorithmic}
\usepackage{algorithm}
\usepackage{array}
\usepackage[caption=false,font=normalsize,labelfont=sf,textfont=sf]{subfig}
\usepackage{textcomp}
\usepackage{stfloats}
\usepackage{url}
\usepackage{verbatim}
\usepackage{graphicx}
\usepackage{booktabs} % better hlines for tables
\usepackage{cite}
\usepackage[dvipsnames]{xcolor}
\usepackage{lipsum}

\makeatletter
\let\NAT@parse\undefined
\makeatother
\usepackage{hyperref}

\hypersetup{
    colorlinks,
    linkcolor={red!50!black},
    citecolor={red!50!black},
    urlcolor={blue!80!black}
}
\newcommand{\APAll}[0]{mAP\textsubscript{2,4,8}\;}
\newcommand{\APAllArrow}[0]{\APAll($\uparrow$)}

\newcommand{\AKDArrow}[0]{AKD($\downarrow$)}

\newcommand{\pp}[0]{\textit{pp} }
\IEEEoverridecommandlockouts    % This command is only needed if you want to use the \thanks command
\title{\LARGE \bf{Learning Keypoints for Robotic Cloth Manipulation  using Synthetic Data}}

\author{Thomas Lips$^{1}$, Victor-Louis De Gusseme$^{1}$ and  Francis wyffels$^{1}$
\thanks{$^{1}$AI and Robotics Lab, Ghent University - imec \newline Technologiepark 126, 9052 Zwijnaarde, Belgium}
\thanks{corresponding author: Thomas.Lips@UGent.be}
}

\begin{document}

\maketitle
\thispagestyle{empty} % change to empty to discard page numbers (official conf template)
\pagestyle{empty}
\begin{abstract}
% context
Assistive robots should be able to wash, fold or iron clothes.
% need
However, due to the variety, deformability and self-occlusions of clothes, creating robot systems for cloth manipulation is challenging. Synthetic data is a promising direction to improve generalization, but the sim-to-real gap limits its effectiveness. 
% Task
To advance the use of synthetic data for cloth manipulation tasks such as robotic folding, we present a synthetic data pipeline to train keypoint detectors for almost-flattened cloth items. To evaluate its performance, we have also collected a real-world dataset.

% Findings
We train detectors for both T-shirts, towels and shorts and obtain an average precision of 64\% and an average keypoint distance of 18 pixels. Fine-tuning on real-world data improves performance to 74\% mAP and an average distance of only 9 pixels. Furthermore, we describe failure modes of the keypoint detectors and compare different approaches to obtain cloth meshes and materials.
% Conclusion
We also quantify the remaining sim-to-real gap and argue that further improvements to the fidelity of cloth assets will be required to further reduce this gap. The code, dataset and trained models are available \href{https://github.com/tlpss/synthetic-cloth-data}{here}.

\end{abstract}


\section{Introduction}
Clothes are everywhere in our living environments. Therefore assistive robots should be able to interact with cloth items and perform tasks such as washing, folding or ironing.  However, due to their variety in shape and appearance, complex deformations and self-occlusions, cloth manipulation and perception are challenging~\cite{jimenez2017visual,yin2021deformablereview2}. Progress has been made in recent years thanks to the continued integration of data-driven techniques~\cite{ha2022flingbot,avigal2022speedfolding,canberk2023clothfunnels} and the development of new hardware~\cite{proesmans2023unfoldir}, but we do not yet have robots that can reliably flatten, fold or iron arbitrary pieces of cloth. 

Part of the reason is that we lack the required amount of data to make these systems generalize. The use of synthetic data has proven an effective direction to overcome this data limit in various domains of robotics~\cite{tobin2017domainrandomization,tremblay2018sim-mix,rudin2022QuadpedIsaac} and beyond~\cite{wood2021fakeit}. Synthetic data can provide arbitrary amounts of perfectly annotated data, but the main concern is to bridge the gap between synthetic and real data, as perfectly covering the real data distribution is often infeasible. Various approaches have been explored to overcome this~\cite{tobin2017domainrandomization,zhu2017cyclegan,tremblay2018sim-mix,wood2021fakeit}, yet a great deal of engineering effort is usually involved in making sim-to-real work. 

In this paper, we want to enable generic robot folding. To this end, we generate synthetic data to learn keypoints on various cloth items after they haven been unfolded by a robot. Current unfolding systems are not perfect yet~\cite{ha2022flingbot,canberk2023clothfunnels,proesmans2023unfoldir} and we therefore detect keypoints on both flattened and slightly deformed clothes. We refer to these cloth states as \textit{almost-flattened}. We detect non-occluded keypoints on clothes of various categories from an RGB image. Using these keypoints, tasks such as folding flattened clothes can be achieved with scripted motions~\cite{doumanoglou2016folding,de2022effective,canberk2023clothfunnels}. Several prior works use synthetic data for cloth manipulation~\cite{corona2018clothpoints,seita2020smoothing,canberk2023clothfunnels} but to the best of our knowledge, none have tackled this specific setting. Furthermore, we are also the first to compare the performance of different procedures to obtain cloth meshes and materials.

\begin{figure}
    \centering
    \includegraphics[width=0.99\linewidth]{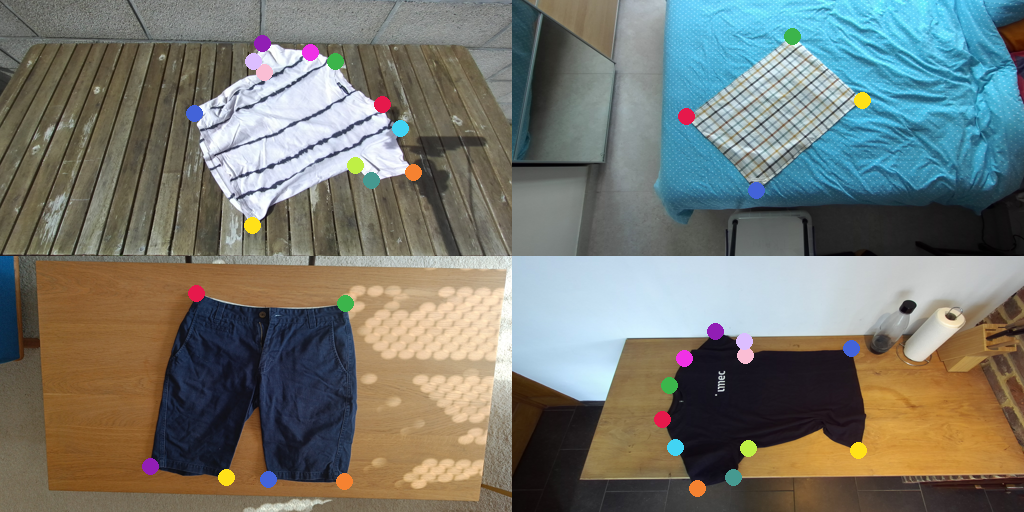}
    \caption{In this work we learn to detect semantic keypoints on \textit{almost-flattened} clothes in everyday environments. To tackle the large diversity in cloth states, cloth materials and environments, we generate synthetic data to train these keypoint detectors. }
    \label{fig:kp-good-examples}
\end{figure}

We generate synthetic data for T-shirts, towels and shorts. This process has three stages: We first procedurally create cloth meshes. In the second stage, these meshes are dropped and deformed to mimic the output of current robotic unfolding systems, using Nvidia Flex~\cite{NvidiaFlex}. Finally, we render images of the clothes and generate the corresponding annotations. To provide real-world training data and to thoroughly evaluate the performance of the keypoint detectors, we have also gathered a dataset of almost 2,000 images.

We find that keypoint detectors trained on our synthetic data have a mean average precision of 64.3 \% on our evaluation dataset, which is an improvement over the baseline trained only on real data. Fine-tuning using real data further improves performance to 74.2\%. These results show that our synthetic data pipeline can be used to build state estimators for robotic cloth folding systems. Additionally, we compare several different procedures to generate cloth meshes and materials where we find that the best results are obtained by using random materials and single-layer cloth meshes, even though this decreases fidelity. Finally, we quantify the remaining sim-to-real gap and argue that this gap can only be closed by further advancements in cloth asset generation and simulation.

To summarize, our contributions are as follows:
\begin{itemize}

    \item We build a synthetic data pipeline and use it to train keypoint detectors that enable cloth folding.
    \item We compare different procedures to obtain cloth meshes and materials, to gain more insight into synthetic data generation for cloth manipulation.
    \item We provide a real-world dataset of 2,000 images and corresponding annotations of almost-flatted clothes in everyday environments.
\end{itemize}

\section{Related Work}

\subsection{Robotic Cloth Manipulation}

Robotic cloth manipulation has been extensively studied~\cite{jimenez2017visual,zhu2022deformablereview1,doumanoglou2016folding,avigal2022speedfolding,ha2022flingbot,de2022effective}. Most works in robotic laundering focus on unfolding (also called flattening) and folding. Despite the considerable success demonstrated by unfolding pipelines~\cite{canberk2023clothfunnels,proesmans2023unfoldir,avigal2022speedfolding}, they frequently yield clothes that aren't perfectly flattened and have yet to fully demonstrate the desired generalization across various cloth instances and environments

Given flattened clothes, tasks such as folding can be tackled using an appropriate representation and a scripted policy~\cite{canberk2023clothfunnels,de2022effective}. Several methods have been explored to create such state representations for clothes, including template fitting~\cite{doumanoglou2016folding,avigal2022speedfolding}, edge detection~\cite{qian2020clothsegmentation} and semantic keypoint detection~\cite{lips2022learning,canberk2023clothfunnels,corona2018clothpoints,seita2019keypoints}. Many of these works use depth images~\cite{corona2018clothpoints,qian2020clothsegmentation,seita2019keypoints}, as these only represent geometry and are not influenced by irrelevant environment elements such as lighting. However, depth images also discard useful information (such as color patterns, seams..) and can lack the required precision~\cite{qian2020clothsegmentation}. To avoid these limitations, we use RGB images in this work.

In this work, we detect keypoints of almost-flattened clothes that are lying on a surface. By doing so, we aim to facilitate the creation of cloth manipulation systems that operate on the output of an unfolding system and can generalize to arbitrary clothes and environments. 

\subsection{Synthetic data for Robotic Cloth Manipulation}
Synthetic data has been used extensively in robotic cloth manipulation research to learn representations and end-to-end policies~\cite{matas2018sim2real,corona2018clothpoints,seita2020smoothing,canberk2023clothfunnels,lips2022learning}.
A variety of cloth simulators have been used, including Pybullet~\cite{Pybullet}, Nvidia Flex\cite{NvidiaFlex} and  Blender~\cite{Blender}. 

As always when learning from synthetic data, obtaining appropriate 3D assets is a key challenge. Many works use a limited number of pre-made cloth meshes which they manually annotate if needed~\cite{corona2018clothpoints,canberk2023clothfunnels,ganapathi2021DON}. Others generate single-layer meshes procedurally~\cite{matas2018sim2real,lips2022learning}. The authors of \cite{bertiche2020cloth3d} have focused on generating a large dataset of cloth meshes, which is used in~\cite{canberk2023clothfunnels,ha2022flingbot}, though these meshes do not come with annotations of semantic locations or edges.

%Many of these works generate depth images\cite{corona2018clothpoints,zhang2020clothkeypoints} but those that generate RGB images mostly use uniform colors as cloth material~\cite{lips2022learning,ganapathi2021DON,canberk2023clothfunnels}, though some also generate normal maps to better mimic real fabric~\cite{lips2022learning,ganapathi2021DON}.

In this work, we further explore the use of synthetic data for keypoint detection. Our goal is to create a pipeline from which models can be trained that generalize to arbitrary clothes, deformations and environments.
%We build a data generation pipeline for different cloth categories and thoroughly evaluate its generalization performance.

\section{ almost-ready-to-fold Clothes Dataset}
\begin{table}
    \vspace {7pt}
    \centering
    \caption{The aRTF Clothes dataset in numbers.  Both scenes and cloth items are distinct for the train and test splits to measure the generalization performance of trained models.
}
    \label{tab:artf-numbers}
    \begin{tabular}{lcccccc}
        \toprule
        &
        \multicolumn{2}{c}{\textbf{\# scenes}} &
        \multicolumn{2}{c}{\textbf{\# cloth items}} &
        \multicolumn{2}{c}{\textbf{\# Images}} \\
         \textbf{Cloth Category} & Train & Test & Train & Test & Train & Test \\ 
        \midrule
         Towels&6  & 8 &  15 & 20 & 210 & 400\\
         T-shirts & 6 & 8 & 15 & 20 & 210 & 400 \\
         Shorts & 6 & 8 & 8 & 9 & 112 & 180 \\ 
         Boxershorts & 6 & 8 & 11 & 11 & 154 & 220 \\
    
        \midrule
        Total&  6 & 8 & 49 & 60 & 686 & 1200\\
        \bottomrule
    \end{tabular}

\end{table}
\label{section:artf-dataset}
% also state that there do not exist such datasets?
To train and evaluate the keypoint detectors, we have created a dataset of almost-flattened clothes in which we mimic the output of current robotic unfolding systems~\cite{canberk2023clothfunnels,proesmans2023unfoldir,ha2022flingbot} by applying small deformations to the clothes. We refer to this dataset as the aRTF Clothes dataset, for almost-ready-to-fold clothes. There have already been some efforts to collect datasets for robotic cloth manipulation~\cite{avigal2022speedfolding,verleysen2020video,ziegler2020fashion}, but to the best of our knowledge, this dataset is the first to provide annotated real-world of almost-flattened cloth items in household settings.

In the next sections, we briefly describe the data capturing and labeling procedure.

\subsection{Dataset Capturing}
We collected data in 14 distinct household scenes. The cloth items were provided by lab members, though we included towels from the household cloth dataset~\cite{garcia2022household}.
% scene collection
%To collect the images we listed 14 household scenes containing a flat surface that can be used to fold clothes.
% cloth collection
%We then collected several clothing items by asking lab members to bring them. For the towels, we included some pieces from the household cloth dataset~\cite{garcia2022household}.

% cloth positioning protocol & deformation protocol
For each image, we started by laying the cloth item on the folding surface and then randomly applied some deformations.
The deformations include a random pinch to create wrinkles, or folding a side or corner of the cloth upwards or downwards and were determined by listing failure modes of state-of-the-art unfolding pipelines~\cite{avigal2022speedfolding,canberk2023clothfunnels,proesmans2023unfoldir}.

% cameras
We collected the images using a ZED2i RGB-D camera and a smartphone.
%For a number of scenes we used a smartphone alongside the ZED camera, to incorporate camera and lens diversity.
The camera was positioned to mimic an ego-centric perspective at a distance between 0.5 and 1\,m.

The dataset contains a total of 1,896 images and is summarized in Table~\ref{tab:artf-numbers}. Collecting these images took 2 persons about 15 hours. Fig.~\ref{fig:kp-good-examples} shows some examples from the dataset.

\subsection{Annotating}   
For each cloth category, we defined a number of semantic locations (keypoints) and annotated these on each image. We use the same semantic locations as Miller et al.~\cite{miller2011parametrized}, except for a single semantic location in the neck opening of T-shirts that we have not used in our work. The semantic locations can be seen in the examples in Fig.~\ref{fig:kp-good-examples}. 

% To avoid symmetries in the annotations, we manually assign an order to the towel keypoints: We take the first corner as the one closest to the top left corner of the bounding box. The second corner is the neighbor of the first one, closest to the top right corner of the bounding box. This determines the order of the other corners as well.
% %% symmetries
% T-shirts, shorts and boxershorts have no symmetries, though we note that in deformed states we often had to use prior knowledge of the specific cloth item to distinguish front and back. For training the keypoint detectors we have therefore considered this as a symmetry as well, as discussed later. 
% segmentation masks
We also labeled the segmentation mask and bounding boxes and use Segment Anything (SAM)~\cite{kirillov2023segment} for pre-labeling the masks. 
%We have found SAM regularly fails with a single point, but produces satisfactory masks when provided with or 2 to 3 points on the cloth surface.
In total, this labeling effort took a single person about 3 days.

\section{Procedural Data Generation Pipeline}
\begin{figure}
    \vspace {7pt}
    \centering
    \includegraphics[width=0.99\linewidth]{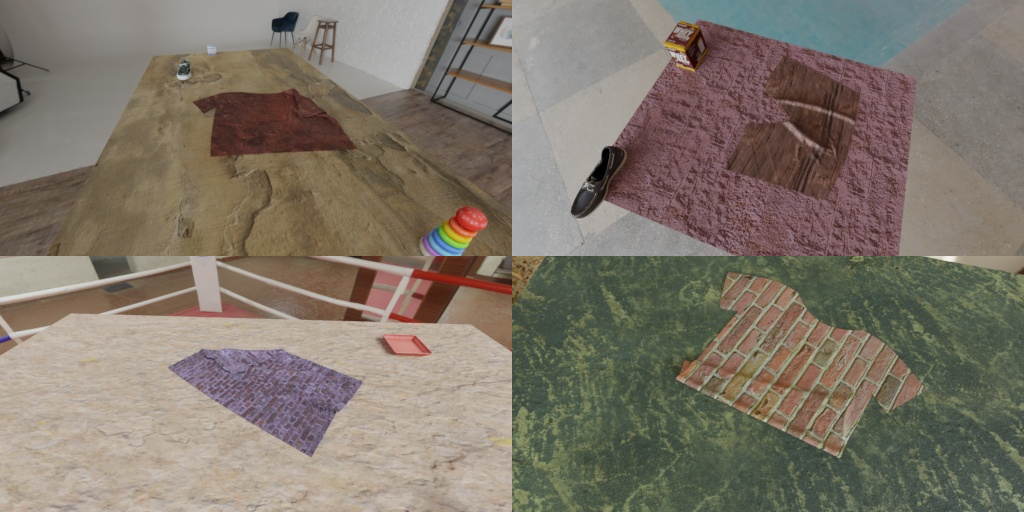}
    \caption{Examples of the generated synthetic images. Though these single-layer meshes with random materials look unrealistic, they have the best performance out of all evaluated procedures.}
    \label{fig:enter-label}
\end{figure}
\label{section:data-generation-pipeline}
In this section, we describe the synthetic data pipeline we have created to generate synthetic images of clothes. In general, generating synthetic data entails two steps: obtaining the assets (meshes, materials...) and specifying how these assets should be used to create 3D scenes, from which the desired images and their corresponding annotations can then be generated. Tools such as Blender\cite{Blender} have enabled the latter step. The main challenge typically lies in gathering or creating the required assets, and this typically comes with a trade-off between realism and diversity. As we are dealing with deformable objects, generating the assets is even more complicated: we need not only to obtain cloth meshes but also to generate the desired distribution of configurations for them, i.e., we need to deform them. The next sections will discuss in more detail how we generate cloth meshes and materials, deform the clothes using Nvidia Flex~\cite{NvidiaFlex}, and use Blender~\cite{Blender} to generate images and the corresponding annotations.

\subsection{Obtaining Cloth Assets}

\subsubsection{Cloth meshes}
\label{section:data-gen-cloth-mesh-procedure}
The first step is to obtain flat cloth meshes. We generate a series of 2D boundary vertices using a template for each cloth type. These templates are inspired on~\cite{miller2011parametrized}. 
We then connect the vertices using bezier curves to create single-layer meshes. These bezier curves are used to create the neck of a T-shirt and to better mimic real cloth items by slightly curving edges. For the same reason, we also round all corners of the single-layer mesh. The parameters of the skeleton, the bezier curves and the rounding radii for all corners are sampled from manually tuned ranges. Finally, the meshes are triangulated such that they have edge lengths of at most 1cm and UV maps are generated. We keep track of the vertices that correspond to each semantic location to automatically label the keypoints later on.

\subsubsection{Cloth Materials}
Faithfully reproducing cloth materials for rendering requires both mimicking fabrics and producing realistic color patterns. At the same time, we also need these materials to provide sufficient diversity to enable the generalization of the models trained on the data. This combination tends to be infeasible. We have experimented with various procedures (see Section \ref{section:experiments-cloth-mesh}) to generate the cloth materials and have found that using random textures from PolyHaven~\cite{PolyHaven} provides the best results. To increase diversity, we mix these textures with a random color.

\subsection{Deforming the Meshes}
\label{section:data-gen-mesh-deformation}
To generate diverse mesh configurations from the flattened meshes, we deform them using the Nvidia Flex cloth simulator~\cite{NvidiaFlex}, which is also used in several previous works on robotic cloth manipulation~\cite{lin2021softgym,ha2022flingbot,canberk2023clothfunnels}. We have also experimented with the Blender cloth simulator~\cite{Blender} and Pybullet~\cite{Pybullet}, but found Nvidia Flex to perform better.

To deform the meshes, we randomize their orientation and drop them on a surface to create wrinkles. Afterwards, we sometimes grasp a point on the cloth and use a circular trajectory to create folds. To create both visible and invisible folds, we sometimes flip the cloth by lifting it entirely, rotating it and dropping it again. This procedure differs from the drop-pick-drop procedure used in~\cite{canberk2023clothfunnels,seita2020smoothing}, as we focus on \textit{almost-flattened} clothes instead of arbitrary crumpled states. The parameters of the procedure have been tuned manually to maximize performance.

To load a mesh into Nvidia Flex, we convert all vertices of the mesh to particles and add both stretch stiffness and bending stiffness constraints using the 1-ring and 2-ring neighbors of each vertex, similar to~\cite{canberk2023clothfunnels,lin2021softgym}.  
%We have also found it very important for the stability of Nvidia Flex to set the rest distance (which determines the cloth thickness) to a value close to the mesh edge lengths and the particle radius to a value that is only slightly higher.
Additionally, similar to previous work~\cite{ha2022flingbot,corona2018clothpoints,canberk2023clothfunnels}, we randomize various physics parameters, including bending stiffness, stretch stiffness, friction and drag to increase diversity.
% Two important parameters in Flex are the particle radius (sphere of influence of each particle) and the solid rest distance of the particles ('classic particle system radius'). This rest distance will also determine the thickness of the cloth when multiple layers fall on each other. 

% \begin{table*}
%     \vspace {7pt}

%     \centering
%     \caption{Performance of the keypoint detector for various data sources. Training on synthetic data outperforms the baseline trained only on real data. The performance is significantly improved by training on synthetic data and then fine-tuning on real data.}
%     \label{tab:main-results}
%     \begin{tabular}{lcccccc}
%         \toprule
%          & \multicolumn{3}{c}{\textbf{\APAllArrow}} & \multicolumn{3}{c}{\textbf{\APTwoArrow}} \\
%         \cmidrule(lr){2-4}
%         \cmidrule(lr){5-7}
%        \textbf{Cloth Type} & real-to-real & sim-to-real & (sim+real)-to-real  & real-to-real& sim-to-real & (sim+real)-to-real  \\ 
%         \midrule
%         T-shirt & 54.4 & 58.2 &69.1  &30.5 & 43.8 & 51.9 \\ 
%         Towel & 74.4  & 83.2 & 88.6 & 65.9 & 77.7 & 83.5 \\ 
%         Shorts & 50.7 & 51.4 & 64.9 & 37.5 & 44.0 & 55.7 \\ 
%         \midrule
%         All & 59.8 & 64.3 & 74.2 & 44.6 & 55.2 & 63.7 \\ 
% \bottomrule
%     \end{tabular}
% \end{table*}
\begin{table*}
    \vspace{7pt}
    \centering
    \caption{Performance of the keypoint detector for various data sources. Training on synthetic data outperforms the baseline trained only on real data. The performance is significantly improved by training on synthetic data and then fine-tuning on real data.}
    \label{tab:main-results}
    \begin{tabular}{lcccccc}
        \toprule
         & \multicolumn{3}{c}{\textbf{\APAllArrow}} & \multicolumn{3}{c}{\textbf{\AKDArrow}} \\
        \cmidrule(lr){2-4}
        \cmidrule(lr){5-7}
       \textbf{Cloth Type} & real-to-real & sim-to-real & (sim+real)-to-real  & real-to-real& sim-to-real & (sim+real)-to-real  \\ 
        \midrule
        T-shirt & 54.4 & 58.2 &69.1  &14.2 &14.0  &8.3  \\ 
        Towel & 74.4  & 83.2 & 88.6 & 12.5 & 13.1 & 6.8  \\ 
        Shorts & 50.7 & 51.4 & 64.9 &23.6  &27.3  & 11.2 \\ 
        \midrule
        All & 59.8 & 64.3 & 74.2 & 16.8 & 18.1 & 8.7 \\ 
\bottomrule
    \end{tabular}
\end{table*}
\subsection{Scene Composition}
\label{section:data-gen-images}
% Describe how scenes (and hence data) are created
% \begin{itemize}
%     \item cloth meshes from above
%     \item cloth materials from above
%     \item surface with randomized shape and texture.
%     \item HDRI world (lighting + background)
%     \item distractor objects from GSO
%     \item rendering using Cycles
%     \item annotations
% \end{itemize}

Using the deformed cloth meshes and the cloth materials described before, the procedure to generate a synthetic image and its annotations is explained below.

We sample an environment texture from PolyHaven~\cite{PolyHaven} to create complex scene lighting. At the same time, we use this texture to create a scene background. Thereafter, a rectangular surface is added to the scene to mimic the folding surface. We sample a material from PolyHaven~\cite{PolyHaven} for this plane and randomize its base color to increase diversity and avoid a bias towards certain colors. We then put a cloth mesh on top of the table and solidify the mesh to add some thickness to the layers. Afterwards, we sample a cloth material and apply it to the mesh.
To make the keypoint detectors more robust, we place a number of distractor objects on the surface. We sample the objects from the Google Scanned Objects dataset~\cite{downs2022GSO}.
Next, we randomize the camera position in a spherical cap around the cloth center and point the camera towards the cloth.
At this point we render a single image using Cycles, Blender's physically based renderer, and create a bounding box, segmentation mask and 2D keypoint annotations. For the keypoints, we use raycasting to determine if the vertex that corresponds to the keypoint is occluded. Inspired by~\cite{wood2021fakeit} we have tried to make the synthetic labels resemble the human labels in the aRTF dataset, which are not necessarily similar to the ground truth labels. We consider a keypoint as visible if any vertex in its 2-ring neighborhood is visible in the image, as we empirically observed this to correspond better to our human annotations.

\subsection{Dataset Generation}
The 3-stage procedure explained in Sections~\ref{section:data-gen-mesh-deformation}, \ref{section:data-gen-mesh-deformation} and~ \ref{section:data-gen-images} is used to generate the desired amount of images for all cloth categories. Creating a mesh template is very fast. Deforming a mesh takes about 6 seconds. Rendering takes about 5 seconds for a 512x256 image on our workstation with an Nvidia RTX 3090, 32GB of RAM and an Intel i7 CPU. More details on the procedural data generation pipeline can be found in the accompanying  \href{https://github.com/tlpss/synthetic-cloth-data}{codebase}.

\section{Keypoint Detection}
\label{section:keypoint-detection-method}
This section provides more details about the keypoint detectors. We describe the model used to detect keypoints and the metrics we use to quantify the performance of the keypoint detectors. We also describe how we ordered the keypoints in image space to deal with the symmetries of cloth objects.

\subsection{Model and Training}
We detect the keypoints on the full-resolution images and formulate keypoint detection as regression on 2D heatmaps as in~\cite{zhou2019objects-as-points}. We convert keypoint coordinates into Gaussian blobs on a 2D probability heatmap, and then use these heatmaps (one for each keypoint category) as target for pixel-wise logistic regression using a binary cross entropy loss. The keypoint model uses a Unet~\cite{ronneberger2015u} inspired architecture and has N full-resolution heatmaps as output. The encoder of the Unet is a pretrained MaxViT~\cite{tu2022maxvit} \textit{nano} model, obtained from the timm library~\cite{rw2019timm}. Keypoints are extracted from the predicted heatmaps by searching for local maxima in a 3x3 pixel grid with a threshold of 0.01 on the probability. Downstream applications can then select the keypoint with the highest probability or use the multi-modal heatmaps to reason more globally over the predictions.

As is common ~\cite{wood2021fakeit,tremblay2018sim-mix}, we employ a set of image augmentations during training: random cropping, random Gaussian blurring and noise, and random color, contrast and brightness augmentations.

%Note that some of these effects could be added to the synthetic data pipeline, but we follow~\cite{wood2021fakeit} by adding them as augmentations to increase flexibility.

We train models on synthetic and/or the train split of the aRTF dataset (see Table~\ref{tab:artf-numbers}). As the synthetic datasets tend to have an order of magnitude more data compared to our aRTF dataset, we use a different set of hyperparameters for each setting. These were obtained using a random hyperparameter search. 

We train each model until convergence and use a subset of the aRTF train split as validation set for this purpose. The performance of each model is tested on the test split of the aRTF dataset, using the checkpoint with the best validation performance. 
% \begin{itemize}
%     \item heatmap-regression based approach as in  [S3K, Jakab, My own previous paper]. Keypoints are converted into a gaussian heatmap onto their category-specific output channel. Using a pixel-wise BCE loss, the network learns to predict these heatmaps, from which local maxima can then be extracted as keypoint predictions.
%     As we are focused on robotic manipulation, we only predict the visible keypoints as these provide reachable grasp poses.
    
%     \item the model uses a U-net inspired architecture with down- and upsampling layers as well as skip connections. For the encoder of the Unet, we use a pretrained MaxViT [X] model, which we have found to work better than a.o. ConvNext models. We use the nano model from the timm library.
%     %\item 3 seeds for each dataset to eliminate its influence.
%     \item We use a different set of hyperparameters for the sim2real and real2real experiments. The hyperparameters were obtained with a random hparam search.
%     \item We use a subset of the aRTF dataset train split as validation set and train all models until convergence.
%     \item We test with the checkpoint that had the highest validation performance.

% \end{itemize}
\subsection{Evaluation Metrics}
\textbf{Mean Average Precision.}
In this work we want to enable folding and hence we only detect keypoints that are not occluded, as a robot can only interact with those parts of the cloth. This implies a varying number of detections for each keypoint category. Furthermore our model can predict an arbitrary amount of keypoints, as discussed in Section~\ref{section:keypoint-detection-method}. Therefore, we use the Average Precision (AP) as in COCO~\cite{lin2014coco} to measure the performance of our models. For each heatmap channel, we classify all predictions as false or true positives using the L2 distance to the ground truth keypoints with thresholds  of 2, 4 and 8 pixels. %For the ZED 2i with an image resolution of 512x256 and given that the maximal distance between a cloth and the camera in the aRTF dataset is 1 meter (see Section~\ref{section:artf-dataset})
For the aRTF dataset and our image resolution of 512 by 256 pixels, this corresponds to real-world distances between real and predicted keypoints of at most 1, 2 and 4\,cm respectively.
%The thresholds were chosen as follows: 1cm and 2cm are relevant for grasping as they are the half-width of regular fingertips and of custom fingertips used in~\cite{lips2022learning}. 4cm is relevant for tasks such as ironing and serves as upper bound to distinguish between lack of precision and lack of semantic knowledge for the keypoint detectors. 
We report the mean average precision over all thresholds (\APAll).
Note that unlike in COCO's object-keypoint similarity (OKS),  we do not scale the distance according to the object size as we believe that absolute errors are more appropriate for robotic manipulation: it does not matter if a T-shirt is a large or small, if you want to grasp it at a specific location you need the same accuracy for both.

\textbf{Average keypoint Distance.}
Next to this detection measure, we also employ a distance-based metric to quantify performance  as in~\cite{seita2019keypoints, corona2018clothpoints}.
More specifically, we report the average keypoint distance (AKS), measured in pixels, between the predicted (if any) keypoint with the highest probability and the non-occluded ground truth keypoints.

%\thomas{Should we discuss that these are only proxy for robot performance that allow for faster/better development?}

%Note that these metrics are more strict than those used in previous works [Corona: 2-4cm as most strict,]

% robotic sucess rate is the ultimate measure, should I discuss why I don't use it here?

\begin{figure*}
    \vspace{7pt}
    \centering
    \includegraphics[width=0.99\linewidth]{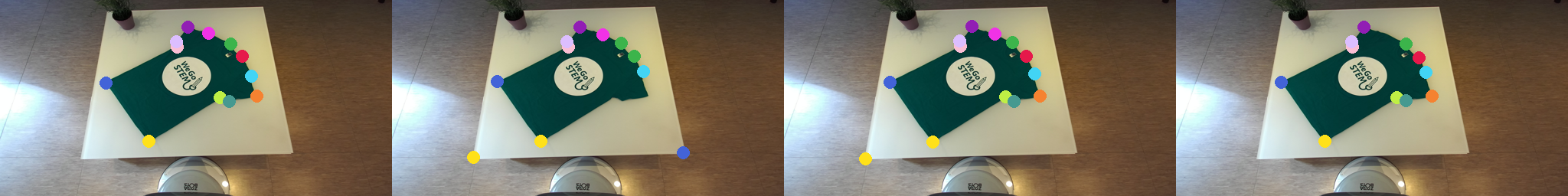}
    \includegraphics[width=0.99\linewidth]{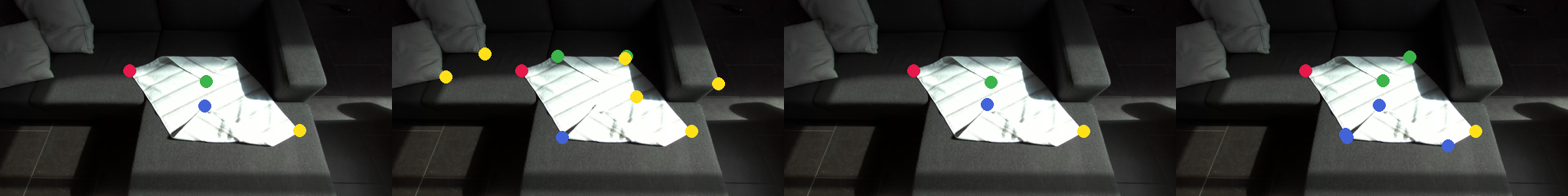}
    \includegraphics[width=0.99\linewidth]{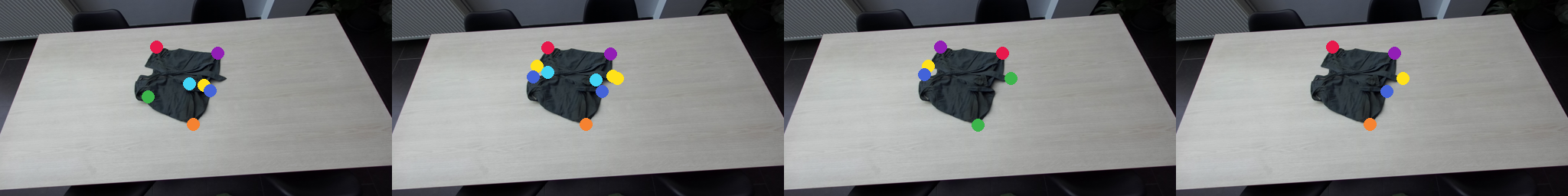}
    \caption{Illustration of failure modes of the detectors. From left to right, the columns show ground truth keypoints and all predicted keypoints of detectors trained on real data, synthetic data and on both. The first row illustrates how the real baseline produces incomplete and inconsistent keypoints, whereas the second row shows how the detectors still struggle with folds. In the third row, the detectors are confused by an open zipper and mistake it for a leg of the shorts. The keypoint colors encode their category for each cloth type.}
    \label{fig:kp-failure-examples}
\end{figure*}

\subsection{Dealing with symmetry}
Clothes have (quasi-) symmetries, which means they can have the same appearance on an image for different configurations (e.g. rotating a white towel on a table by 180 degrees will result in the same image). These symmetries must be handled carefully because the keypoint detector cannot distinguish between them. 

One possible solution is to not distinguish between symmetric keypoints by labeling them identically~\cite{lips2022learning},  but this results in some loss of information: we can no longer infer the connectivity of the keypoints, which is useful if you want to e.g. grasp the two corners of an edge. Other solutions include using a loss function that is invariant across all symmetries~\cite{corona2018clothpoints}, or selecting an arbitrary ordering in the image~\cite{canberk2023clothfunnels,ganapathi2021DON}. A thorough evaluation of these approaches would be out of scope and in this work we have applied an ordering to the keypoints to deal with symmetry. To do so, we have adapted the labels of the synthetic data and the aRTF dataset as follows: 

For towels, we have taken the keypoint closest to the top left corner of the bounding box to be the first and its physically adjacent corner that is closest to the top right corner of the bounding box as the second keypoint. The others follow from this convention.

T-shirts or shorts have no rigid motions symmetries. However, detecting the front/back side can be very ambiguous, even on real data, and therefore we treat this as a rotational symmetry. For T-shirts we take as \textit{left} side, the side of which the waist keypoint is closest to the bottom left corner of the T-shirt's bounding box. For shorts, we use the top left corner of the bounding box. 

\section{Experiments}
In this section, we train several keypoint detectors on different data sources. We evaluate the performance of all models on the test split of the aRTF dataset (see Table~\ref{tab:artf-numbers}) using the mean Average Precision (mAP) as metric. In Section~\ref{section:experiments-main} we show that training on the synthetic data (sim-to-real) for three cloth categories results in better performance than training on the aRTF train split (real-to-real), and how fine-tuning on the real data (sim+real-to-real) improves performance significantly. The cloth mesh procedure and cloth materials described in Section~\ref{section:data-generation-pipeline} are compared against other alternatives in Sections~\ref{section:experiments-cloth-mesh} and~\ref{section:experiment-cloth-material}.

\subsection{Main Results}

\begin{table}
    \centering
    \caption{reality gap of the synthetic data as measured by the difference between sim-to-sim and sim-to-real performance.}
    \label{tab:reality-gap}
\begin{tabular}{lcccc} 
\toprule 
 & \multicolumn{2}{c}{\textbf{\APAllArrow}} & \multicolumn{2}{c}{\textbf{\AKDArrow}} \\
\cmidrule(lr){2-3} \cmidrule(lr){4-5}
\textbf{cloth type} & sim-to-real & sim-to-sim  & sim-to-real & sim-to-sim  \\
\midrule
T-shirt & 58.2 & 81.8 & 14.0 & 5.8 \\
Towel & 83.2 & 84.7 & 13.1 & 7.4 \\
Shorts & 51.4 & 79.8 & 27.3 & 7.4 \\
\midrule
All & 64.3 & 82.1 & 18.1 & 6.8 \\
\bottomrule
\end{tabular}
\end{table}

% sim2real slightly better than real2real, for 3 categories.
% finetuning on real results in better performance, also for 3 categories.
% can be noted taht towels perform better across the board (20% better for mAPALL, 30% for mAP2)!

% examples of succesful predictions 

% qualitative analysis of what goes wrong: show some examples as well. 
% - defomrations
% - non consistent
% ... 

% there is still a big reality gap: difference between sim-to-sim and sim-to-real

\label{section:experiments-main}
% \begin{itemize}
%     \item performance of sim2real, compare to baseline.
%     \item additional performance when finetuning, this denotes a sim2real gap. Can also be seen when comparing sim2sim to sim2real (separate table).
%     \item differences between categories? towels a lot better
%     \item some example predictions that go well.
%     \item some that go wrong (folds vs edges when inwards)
%     \item all struggle with folds
%     \item ordering: predicts one wrong and all others can be off?
%     \item consistency: can sometimes predict combinations that do not seem self-consistent!
    
% \end{itemize}
We have trained keypoint detectors for both T-shirts, shorts and towels. For each cloth type, we trained three models using different data sources: the corresponding train split of the aRTF dataset (which we use as a baseline), a dataset of synthetic images and a combination of both by fine-tuning on the aRTF data after training on the synthetic data. We use 10,000 synthetic images per category as we have observed that surpassing this amount does not lead to a significant improvement. 

The performance of all models, as measured on the aRTF test splits, is given in Table~\ref{tab:main-results}. Over all cloth types, the baseline model that is only trained on real data obtains an \APAll of $59.8$ and an AKD of $16.8$ pixels.  Training on synthetic data results in an improved \APAll ($64.3$) but a lower AKD of $18.1$ pixels, caused by a longer tail of the keypoint distances for the model trained on synthetic data.  Fine-tuning on the aRTF data results in an \APAll of $74.2$ and an AKD of only $8.7$ pixels, almost halving the average distance compared to training on real or synthetic data only. When comparing between categories the performance is best for the towels on both metrics.
A few examples of the output of the models trained on a combination of the aRTF train split and synthetic data can be found in Fig.~\ref{fig:kp-good-examples}. We have also added a \href{https://youtu.be/Mlwg_qPxr78}{video} to the repository where we visualize the keypoint predictions while interacting with various clothes. Using the keypoint detectors, folding policies can be scripted~\cite{de2022effective,canberk2023clothfunnels}. We demonstrate this by folding T-shirts using scripted motions based on the detected keypoints. We do not report quantitative results, for which our experimerent was not extensive enough, but have included a \href{https://youtu.be/bqnQ4iLnp20}{video} in the project repository.

The above results clearly show the added value of the synthetic data. At the same time, Table~\ref{tab:main-results} shows that the detectors are not perfect. To provide more insight into the failures of the detectors, we show the predicted keypoints of the models trained on each data source on a number of samples from the aRTF test split in Fig.~\ref{fig:kp-failure-examples}. Each example illustrates a common failure mode: The first row illustrates how training only on the aRTF train split results in models that confuse semantic locations and return many false positives, which is also visible in the second example. The keypoint detectors still struggle with folds, as can be seen in the second example. Despite many efforts to tailor the distribution of mesh deformations in the synthetic data (see Section~\ref{section:data-gen-mesh-deformation}), we have not managed to completely solve this problem. The final example in Fig.~\ref{fig:kp-failure-examples} illustrates how all models have difficulties with a particular configuration of a short in which the opened zipper is confused for a leg. Similar confusions occur for T-shirts as well. 

%skip to save space
%Even though the keypoints are not detected perfectly on these examples, they do illustrate the increased performance when training on synthetic data and also when further finetuning on real data, as the keypoints tend to get more reliable towards the right.

The lack of further improvement when adding more synthetic data and the performance improvement when fine-tuning on the aRTF data already suggest that there is a reality gap for the synthetic data. To measure this, we have evaluated both the sim-to-sim performance and the sim-to-real performance of the models trained on synthetic data. These results can be found in table~\ref{tab:reality-gap}. Across all categories, the difference is about 20\pp for the mAP and 11 pixels for the AKD, which confirms there is still a substantial reality gap.
%Interestingly, for towels the sim and real performance are very close, suggesting that the gap is smaller for this category. This also corresponds with the results from Table~\ref{tab:main-results}, where the performance increase associated with fine-tuning on the aRTF data was lower than for the other categories.

\subsection{Comparing procedures for obtaining cloth meshes}
\label{section:experiments-cloth-mesh}
\begin{figure}
    \vspace {7pt}

    \centering
    \includegraphics[width=0.99\linewidth]{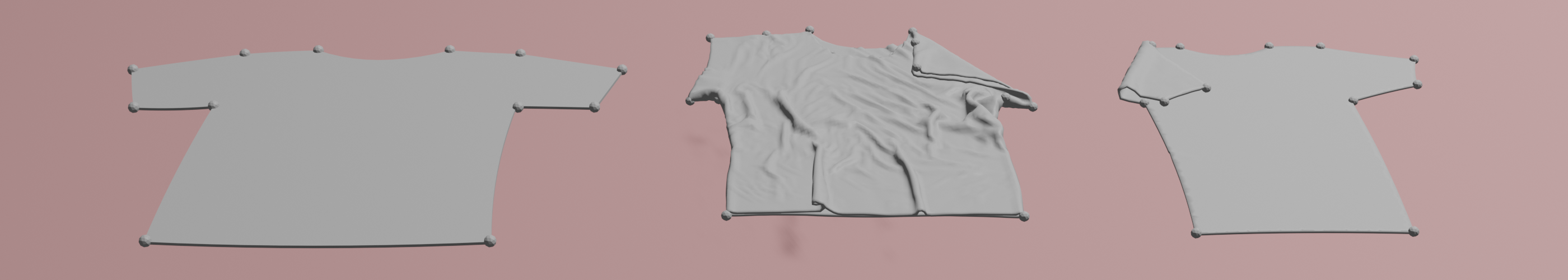}
    \caption{Examples of the considered cloth mesh procedures. Left to right: undeformed single-layer mesh, Cloth3d mesh, single-layer mesh. Though less realistic, using the single-layer meshes results in the best performance.}
    \label{fig:mesh-examples}
\end{figure}

In this experiment, we motivate the use of single-layer meshes for synthetic data generation by comparing three different procedures to generate T-shirt meshes:
\begin{enumerate}
    \item Inspired by~\cite{canberk2023clothfunnels} we have taken 5 T-shirt meshes from the Cloth3D dataset~\cite{bertiche2020cloth3d}. We have triangulated the meshes and cleaned up their UV maps. Using Nvidia Flex, we generated a flattened version of these meshes, by dropping them from a certain height. We then manually labeled all keypoints by selecting the corresponding vertices on the mesh. Similar to~\cite{corona2018clothpoints}, we have created 10 variants of each mesh by rescaling it, leading to a total of 50 different meshes. Using these meshes, we generate deformed configurations with the procedure discussed in Section~\ref{section:data-gen-mesh-deformation}.
    \item We procedurally generate single-layer T-shirt meshes using a 2D template as described in Section~\ref{section:data-gen-cloth-mesh-procedure}. We then create deformations using the same procedure as above. As these meshes and their annotations are generated procedurally, we can generate arbitrary amounts of meshes, unlike for the cloth3d procedure.
    \item We use the same 2D templates as above but do not deform them, which results in perfectly flat meshes, similarly to~\cite{lips2022learning}.
\end{enumerate}
An example of each mesh procedure can be found in Fig.~\ref{fig:mesh-examples}. 
For each procedure, we generate a dataset of 5,000 T-shirt meshes which we then use to generate 5,000 synthetic images using the pipeline described in Section~\ref{section:data-gen-images}. The performance on the aRTF test split of the keypoint detectors trained on these synthetic datasets can be found in table~\ref{tab:mesh-comparison}. Surprisingly, the more realistic meshes from the Cloth3D dataset perform worse than the less realistic but more diverse single-layer meshes. The undeformed single-layer meshes also perform surprisingly well. Based on these results, we use the single-layer mesh procedure in this paper.

\begin{table}[]
    \centering
    \caption{Comparison of different procedures to obtain cloth meshes. The single-layer procedure results in the best performance.}
    \label{tab:mesh-comparison}
    \begin{tabular}{lcc}
        \toprule
        \textbf{Cloth Mesh Procedure} & \textbf{\APAllArrow} & \textbf{\AKDArrow} \\ 
        \midrule
        Cloth3D subset & 43.4  & 20.6 \\
        single-layer & \textbf{54.3} & \textbf{15.3} \\
        single-layer undeformed & 53.8& 17.9\\
        \bottomrule

    \end{tabular}
\end{table}

\subsection{Comparing different cloth materials}
\label{section:experiment-cloth-material}

\begin{figure}
    \vspace {7pt}

    \centering
    \includegraphics[width=0.99\linewidth]{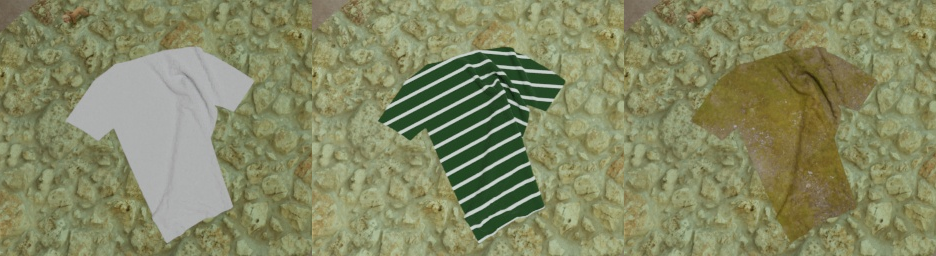}
    %\subcaption{a) uniform  b) cloth-tailored c)  random}
    \caption{Examples of the cloth different materials that were considered. From left to right: uniform colors, a cloth-tailored procedural material and random PolyHaven~\cite{PolyHaven} textures. Though random materials are less plausible, they result in the best performance.}
    \label{fig:material-examples}
\end{figure}

In this experiment, we compare different  cloth material configurations for rendering:
\begin{enumerate}
    \item For the first configuration, we simply add random materials from PolyHaven~\cite{PolyHaven} to the cloth and mix the base color map with a random color to create more diversity.
    \item In the second configuration, we apply a uniform color to the clothes as in~\cite{ha2022flingbot,canberk2023clothfunnels,seita2020smoothing,lips2022learning}.
    \item We also compare against a more tailored procedural material: next to uniform colors, we also add striped color maps. We also add random images to the clothes to mimic logos and prints. Furthermore, we procedurally generate a normal map that mimics fabric materials and generates additional wrinkles, similar to~\cite{lips2022learning,seita2020smoothing}.
\end{enumerate}  
An example of each material can be found in Fig.~\ref{fig:material-examples}. We have generated datasets of 5,000 images for these materials, using the procedure from Section~\ref{section:data-gen-images}. The performance of the keypoint detectors trained on these datasets can be found in table~\ref{tab:material-comparison}. We observe that uniform color materials result in the lowest performance. This is not surprising as the aRTF dataset contains many non-uniform colored cloth items. It is also consistent with findings in our previous work~\cite{lips2022learning}, where we also observed a decreased performance for non-uniform cloth pieces. Perhaps more surprisingly, the random materials perform better than the tailored materials even though the latter look more plausible. We therefore use random materials for all other experiments in this paper.

\begin{table}[]
    \centering
    \caption{Comparison of different Cloth materials. Adding random materials to the cloth meshes leads to the highest performance on the aRTF test split.}
    \label{tab:material-comparison}
    \begin{tabular}{lcc}
        \toprule
        \textbf{Cloth Materials} & \textbf{\APAllArrow} & \textbf{\AKDArrow} \\ 
        \midrule
        PolyHaven random & \textbf{54.3}  & \textbf{15.3} \\
        cloth-tailored  & 49.0 & 22.1\\
        uniform color& 35.1& 35.4 \\
        \bottomrule

    \end{tabular}
\end{table}

\section{Discussion}

% towels > shorts & tshirts. Because towels are easier inherently (this would be surprising as they have less distinctive features and hence more false positives used to occur) Because the images of towels are made easier? Or because there is less label noise? 

% recap 
We have created a synthetic data pipeline for clothes and used it to generate images and corresponding keypoint annotations of cloth items that are \textit{almost} flattened. The improved performance when using this synthetic data to train keypoint detectors clearly shows its benefits. 
% would improve with more real, but never perfect generalization.
At the same time, we stress that the amount of real-world data of the baseline is limited. Scaling up real-world data collection would improve the results of the baseline. However, due to the huge diversity in cloth meshes, their configurations, cloth materials and environments, we argue that synthetic data will still be needed to obtain models that can generalize to all these aspects.

% only way to escape from local optimum is to improve realism.
We have found that performance when training on the synthetic data increased as we traded diversity for fidelity for both cloth meshes and materials. However, we also determined that there is a reality gap that limits the sim-to-real performance. We argue that the current lack of realism in favor of diversity is a local optimum, as we do not expect more randomizations to further improve performance. Further increasing the realism of the data generation pipeline will allow to escape from the observed performance limit.

%% assets 
Improving performance will hence require more accurate cloth meshes that contain fine-grained elements such as seams, zippers, pockets and labels, which are all important cues to determine the semantic locations. To apply materials realistically, the meshes also need to contain accurate UV maps. Furthermore, the meshes also need to be automatically annotated to reduce engineering effort. Leveraging recent advances in generative text-to-3D models~\cite{poole2022dreamfusion} could provide an exciting alternative to manually engineering all the above.

%% physics
Not only do the assets need to be improved, particle-based simulators such as Nvidia Flex~\cite{NvidiaFlex} have fundamental limitations. Next to the limited realism of the cloth physics~\cite{ha2022flingbot,blanco2023benchmarking}, we have also found the inherent intertwining of mesh resolution with cloth thickness and deformation granularity hindering. Simulators such as C-IPC~\cite{li2020cipc} provide improved realism~\cite{de2022effective}, though at the cost of additional computational complexity.
% end-to-end?

Using generative models for end-to-end data augmentation as in \cite{yu2023ROSIE} is an alternative to manual engineering of 3D assets and physics simulators, though we have found in some initial experiments that keeping the labels consistent when augmenting real data is a challenge. 

% deformations still tricky -> how to solve? and really needed?
% we have observed that the keypoint detectors still struggle with folds and severe deformations. This results in predictions that are between themselves not consistent and false positives on the apparent cloth edges. It is very well possible that these issues can be resolved with improved data generation, but we think that the models would benefit from more geometric reasoning capabilities.

Though we firmly believe in the benefits of synthetic data and their use for robotic manipulation, we are not convinced that forcing a model to predict keypoints from a static image of heavily deformed clothes lying on a surface is the best way forward: this might be an artificially hard task. Using more interactive perception as in~\cite{doumanoglou2016folding} would allow the system to collect more information. 
Finally, improvements to the generalization of unfolding systems could also reduce the need to handle deformed pieces in the first place.

% \section{Robot Experiments}
% If needed (@Francis?): can probably report some towel foldings from 'flattened' starting states? But this feels redundant though... other works have already done this?

\section{Conclusion}
In this work, we have described a pipeline to generate synthetic images of clothes with the goal of improving generalization of robotic cloth manipulation systems. By training keypoint detectors on images generated by this pipeline we have validated its usefulness. The resulting models and the aRTF dataset can be used to enable various tasks such as folding or ironing. Furthermore, the pipeline can be easily adapted to train models for other tasks or cloth types.

We have found that models trained on our data generation pipeline, which prioritizes diversity over fidelity, have a decreased performance when transferred to the real world. We argue that future work should focus on increasing fidelity to bridge this reality gap.

A limitation of our work is that we assume the cloth type is already known in advance, which cannot be taken for granted in real-world scenarios. We believe our data generation pipeline can be used for learning to classify clothes as well, but consider this to be out of scope.

%%% Appendices serve to document some additional stuff that I don't want to put in the RA-L paper.
%\input{appendices.tex}

\section*{Acknowledgments}
The authors wish to thank Maxim Bonnaerens and Peter De Roovere for their valuable feedback. This project is supported by the Research Foundation Flanders (FWO) under Grant numbers 1S56022N (TL) and 1SD4421N (VDG) and by the euROBIn Project (EU grant number 101070596).
\bibliographystyle{IEEEtran}
\bibliography{references}

\end{document}